\documentclass{article}
\usepackage{spconf,amsmath,graphicx}

\usepackage{amsfonts}
\usepackage{booktabs}
\usepackage{multirow}
\usepackage{mathrsfs}  
\usepackage{cite}  
\usepackage{hyperref}
\usepackage[nolist]{acronym}
\usepackage{array}

\usepackage{pgfplots}
\usepgfplotslibrary{groupplots}

\usepackage[bottom]{footmisc}

\begin{acronym}
\acro{FR}{Face Recognition}
\acro{YTF}{YouTubeFaces}
\acro{DAN}{Discriminative Aggregation Network}
\acro{TPIR}{True Positive Identification Rate}
\acro{FPIR}{False Positive Identification Rate}
\acro{GAN}{Generative Adversarial Network}
\acro{GAP}{Global Average Pooling}
\acro{LFW}{Labeled Faces in the Wild}
\acro{PLFW}{Partial Labeled Faces in the Wild}
\acro{CE}{Cross-Entropy}
\acro{CNN}{Convolutional Neural Network}
\end{acronym}

\DeclareMathOperator{\softmax}{softmax}
\DeclareMathOperator{\sigmoid}{sigmoid}

\newcommand{\blockb}[3]{\multirow{3}{*}{\(\left[\begin{array}{c}\text{1$\times$1, #2}\\[-.1em] \text{3$\times$3, #2}\\[-.1em] \text{1$\times$1, #1}\end{array}\right]\)$\times$#3}
}

\title{Attention-based Partial Face Recognition}
\name{Stefan H\"ormann \quad Zeyuan Zhang \quad Martin Knoche \quad Torben Teepe \quad Gerhard Rigoll
\thanks{\copyright 2021 IEEE. Personal use of this material is permitted. Permission from IEEE must be obtained for all other uses, in any current or future media, including reprinting/republishing this material for advertising or promotional purposes, creating new collective works, for resale or redistribution to servers or lists, or reuse of any copyrighted component of this work in other works.}}
\address{Technical University of Munich}
\begin{document}
\ninept
\maketitle
\begin{abstract}
Photos of faces captured in unconstrained environments, such as large crowds, still constitute challenges for current face recognition approaches as often faces are occluded by objects or people in the foreground.  However, few studies have addressed the task of recognizing partial faces.  In this paper, we propose a novel approach to partial face recognition capable of recognizing faces with different occluded areas. We achieve this by combining attentional pooling of a ResNet's intermediate feature maps with a separate aggregation module. We further adapt common losses to partial faces in order to ensure that the attention maps are diverse and handle occluded parts. Our thorough analysis demonstrates that we outperform all baselines under multiple benchmark protocols, including naturally and synthetically occluded partial faces. This suggests that our method successfully focuses on the relevant parts of the occluded face.
\end{abstract} %emphasize cross
\section{Introduction}
\label{sec:intro}

\begin{figure}[b]
  \centering
  \vspace{-0.3cm}
\includegraphics[width=0.7\linewidth]{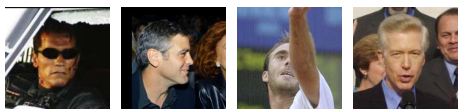}
  \vspace{-0.35cm}
  \caption{Examples of partial faces ocurring in the \ac{LFW} dataset \cite{LFW}.}
  \vspace{-0.15cm}
  \label{fig:examples}
\end{figure}

State-of-the-art \ac{FR} approaches \cite{Liu.2017,wang2018cosface, Arcface, kim2020groupface} achieve satisfying performance under controlled imaging conditions, such as frontal faces, manually aligned images, regular expressions, and consistent illuminations. However, these requirements are often not fulfilled in many practical scenarios due to ineffectual control over the subjects and environments, resulting in partially visible faces. As illustrated in \autoref{fig:examples}, there are multiple examples for partial faces occurring in real-life scenarios: faces with extreme head poses causing face parts to become invisible; faces with intense illuminations or saturations provoking vanishing face details; faces in the background being obstructed by foreground objects or persons; faces at the edge of the image being cut off. 

After face detection, typical \ac{FR} approaches for holistic faces use \acp{CNN} to embed faces into a deep feature space, in which face pairs are considered genuine if their feature distance is lower than a threshold. For partial faces, however, partial face detection algorithms \cite{opitz2016grid, mahbub2016partial, chen2018adversarial} are required to detect face parts even when the face is occluded, yielding a face with an arbitrary resolution. Therefore, partial \ac{FR} approaches need to be capable of either handling arbitrary input resolutions \cite{liao2012partial, hu2013robust, weng2016robust, he2016multiscale, he2018dynamic} or tolerate that the information is only present in a small area within the input \cite{song2019occlusion, xu2020improving}. Moreover, to compare partial with holistic faces, it is desirable to design a network performing well for both faces. 

In our work, we propose an approach to partial \ac{FR} using fixed-size images. The extracted face patches are normalized and zero-padded to match the input resolution. Hence, the face patch is not deformed yet centered, which causes the loss of any spatial information. To compensate for this information loss, we predict attention maps capable of focusing on their respective region of interest independent of their positions. Using attentional pooling followed by aggregation, we obtain a single feature vector robust against the lack of spatial information in face patches.

% Secondly, traditional face recognition CNN models are able to extract features globally perfectly, but rarely have intent on the relevance between different local regions of a face. For partial face verification and identification, it is important that network can realize the connection between two different regions of a face, such as eyes with nose, mouth with eyes. 
%The goal of this thesis is to identify as well as verify a random cropped region from a full face image, which might only contain an eye, or a mouth. Based on those requirements, we propose a novel architecture for partial face recognition. Unlike in \cite{he2018dynamic} they take an arbitrary size of partial face images as input and compare between patches, we use fixed size input images to make use of previous CNN models, for instance, ResNet. The modified ResNet can be used to extract intermediate feature maps. In order to solve the second issue, we also generate some landmark attention maps from the separated third block of ResNet. At the end, the proposed network is capable of not only recognizing full face images, but also recognizing partial faces.

The contributions of our work can be summed up as follows:
\begin{itemize}
  \setlength\itemsep{0.2em}
\item On the example of a ResNet \cite{resnetv2}, we propose an extension that utilizes attentional pooling with an aggregation network and  is trained with two popular losses adapted for partial \ac{FR}.
%\item We propose two modification to popular losses to adapt for partial \ac{FR}.
\item In our exhaustive analysis covering multiple partial \ac{FR} protocols, we show that our modifications substantially improve recognition performance and outperform the baselines for synthetically and naturally occluded partial faces.
\end{itemize}

\begin{figure}[t]
  \centering
\includegraphics[width=0.9\linewidth]{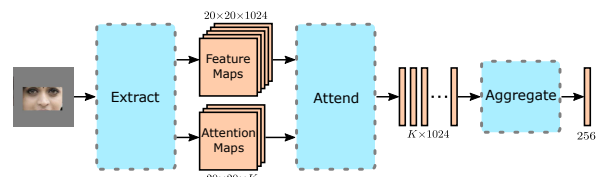}
  \vspace{-0.4cm}
  \caption{Overview of our approach for partial \ac{FR}.}
  \label{fig:partial}
  \vspace{-0.5cm}
\end{figure}
\begin{figure*}[t]
  \centering
\includegraphics[width=1.0\linewidth]{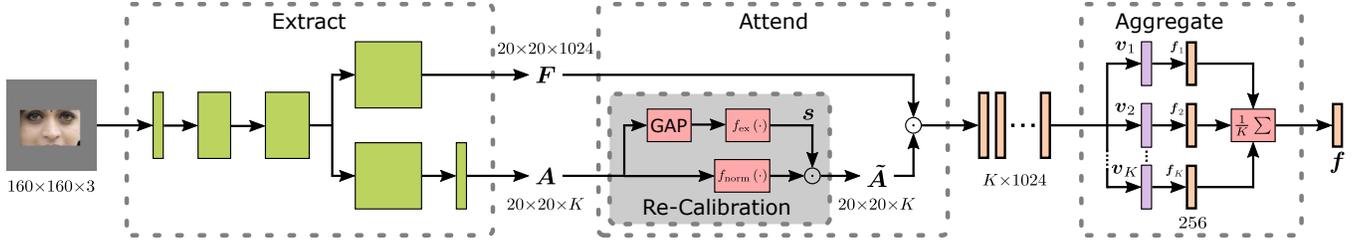}
  \caption{Our proposed partial \ac{FR} approach: A modified ResNet extracts feature maps $\boldsymbol{F}$ and attention maps $\boldsymbol{A}$. The re-calibrated attention maps $\boldsymbol{\tilde{A}}$ are used to pool $\boldsymbol{F}$ into $K$ feature descriptors $\boldsymbol{v}_k$. In the Aggregate module, $K$ independent fully connected layers transform every $\boldsymbol{v}_k$ into a joint feature space, in which the feature vectors $\boldsymbol{f}_k$ are averaged to obtain the final feature $\boldsymbol{f}$.}
  \label{fig:architecture}
\end{figure*}

\section{Related Work}
\label{sec:format}

Traditional partial \ac{FR} approaches can be divided into region-based \cite{sato1998partial,gutta2002investigation,park2010periocular} and keypoint-based \cite{liao2012partial, hu2013robust, weng2016robust} approaches. Region-based approaches extract features from face patches, such as eyes, ears and nose \cite{sato1998partial}, face halves \cite{gutta2002investigation}, or the periocular region \cite{park2010periocular}. Keypoint-based approaches compute descriptors from face patches of arbitrary size. While Liao et al. \cite{liao2012partial} utilized Garbor Ternary Patterns as a descriptor and applied a sparse representation-based classification algorithm, Hu et al. \cite{hu2013robust} focused on SIFT features. Apart from SIFT, Weng et al. \cite{weng2016robust} also incorporated SURF and scale-invariant local binary pattern descriptors.

With emerging deep learning algorithms, He et al. \cite{he2016multiscale} proposed a multiscale region-based \ac{CNN}, which extracts a feature for every face patch at different scales. In order to cope with face patches of arbitrary size, He et al. \cite{he2018dynamic} used dynamic feature learning to match local feature maps, which were obtained by a fully convolutional neural network. However, all previous approaches require overlapping patches during the matching. Thus, a cross-matching of partial faces of, e.g., the eye with the mouth region, is not possible. 

In order to obtain a global face representation and focus on non-occluded face areas, utilizing a siamese network together with a predicted occlusion mask \cite{song2019occlusion} or attention map \cite{xu2020improving} was proposed.

\section{Methodology}
\subsection{Network Architecture}

\autoref{fig:architecture} depicts our partial \ac{FR} approach divided into three modules: Extract, Attend, and Aggregate.
The Extract module extracts feature maps $\boldsymbol{F}\in \mathbb{R}^{20\times20\times1024}$ and attention maps $\boldsymbol{A}\in \mathbb{R}^{20\times20\times K}$ from the input image with $K$ denoting the number of attention maps. In the Attend module, the feature maps are pooled into $K$ intermediate feature vectors using re-calibrated attention maps. The aggregation module maps these intermediate feature vectors into a joint feature space, in which the final feature vector $\boldsymbol{f}\in \mathbb{R}^{256}$ is obtained.

\begin{table}[b!]
\vspace{-0.4cm}
\caption{The architecture of the Extract module. Residual blocks are shown in brackets, with the numbers of blocks stacked.}
\begin{center}
\label{tab:architecture}
\begin{tabular}{cccc}
\toprule
Block & Size & \multicolumn{2}{c}{Layer} \\
\midrule
1 & $80^2$ &\multicolumn{2}{c}{7$\times$7, 64, stride 2}\\
\midrule
\multirow{4}{*}{2} & \multirow{4}{*}{$40^2$} & \multicolumn{2}{c}{3$\times$3 max pool, stride 2} \\
  &  &  \multicolumn{2}{c}{\blockb{256}{\phantom{2}64}{3}} \\
 &  \\
 &   \\
\midrule
\multirow{3}{*}{3} &  \multirow{3}{*}{$20^2$}   & \multicolumn{2}{c}{\blockb{512}{128}{4}}  \\
  &  \\
  &  \\
\midrule
\multirow{4}{*}{4} & \multirow{4}{*}{$20^2$}   & \blockb{1024}{\phantom{1}256}{6} & \blockb{1024}{\phantom{1}256}{6}   \\
  & & & \\
  &  & & \\
  & & & 1$\times$1, $K$\\
\bottomrule
\end{tabular}
\end{center}
\end{table}

\subsubsection{Extract}
Inspired by \cite{xie2018comparator}, we utilize a truncated ResNet-50 architecture \cite{resnetv2}, which is concluded after the fourth block. Hence, we only perform three spatial downsamplings and obtain feature maps of size $20 \times 20$, in which the regions are still well distinguishable. Unlike \cite{xie2018comparator}, we separate the ResNet after the third block to allow both branches to focus on their respective tasks. While we directly obtain $\boldsymbol{F}$ after the fourth ResNet block, we add an extra $1\times1$ convolution with ReLU \cite{nair2010rectified} activation function to obtain $\boldsymbol{A}$. The detailed architecture is summed up in \autoref{tab:architecture}.

The generated attention maps should fulfill the following two key attributes:
1) Attention maps should be mutually exclusive, i.e., different attention maps focus on different regions of a face image; \mbox{2) The attention} maps' activations should  correlate with their respective region's visibility. 

Notably, the implicitly-defined attention map activations do not necessarily follow the same intuition as human-defined facial landmarks such as eyes or nose.% They usually focus on some recognizable parts of a face.

\subsubsection{Attend}

As in \cite{xie2018comparator}, the attention maps $\boldsymbol{A}$ need to  be re-calibrated. Xie et al. \cite{xie2018comparator} proposed the attentional pooling for set-based \ac{FR} normalizing $\boldsymbol{A}$ separately over all images within a set, thereby ensuring that the respective information is extracted from the image with the largest activation in $\boldsymbol{A}$. For partial \ac{FR}, however, we only consider a single image and expect different attention maps to be relevant depending on the region of the face, i.e., if the eyes are occluded, the corresponding attention maps should contain low activations. Thus, we propose to use a parameter-free re-calibration following the structure of  \cite{hu2018squeeze}:

First, we normalize $\boldsymbol{A}$ by applying the sigmoid function $f_{\text{norm}}\left(\cdot\right) = \sigmoid\left(\cdot\right)$. In this way, every pixel in every attention map is normalized separately to $(0;1)$. %However, this normalization dispenses the information 
Besides, we compute a vector $\boldsymbol{s}\in\mathbb{R}^K$ representing the importance of every attention map by applying \ac{GAP} followed by $f_{\text{ex}}\left( \cdot\right)= \softmax\left( \cdot\right)$:
\begin{equation}
\label{eq:s}
\boldsymbol{s} = f_{\text{ex}}\left(\frac{1}{20^2}\sum_{i,j}\boldsymbol{A}_{i,j,k}\right)
\end{equation}
with the indices $i$, $j$, and $k$ denoting the pixel in the $i$-th row and $j$-th column of the $k$-th attention map. By incorporating \ac{GAP}, we obtain global information of all attention maps and transform it into a probability distribution indicating the importance of the respective attention map using the softmax function. Next, we multiply the $k$-th self-normalized attention map $\boldsymbol{A}_k$ with its corresponding importance $s_k$ to obtain the final re-calibrated attention map $\boldsymbol{\tilde{A}}_k$:
\begin{equation}
\boldsymbol{\tilde{A}}_k = s_k\cdot\sigmoid\left(\boldsymbol{A}_k\right)
\end{equation}

Hence, in our re-calibration, we combine local information within each attention map together with global information across the attention maps.

After re-calibration, we apply attentional pooling as in \cite{xie2018comparator} to obtain $K$ feature descriptors $\boldsymbol{v}_k \in \mathbb{R}^{1024}$:
\begin{equation}
\boldsymbol{v}_k = \sum_{i,j}\boldsymbol{F}_{i,j,:}\odot \boldsymbol{\tilde{A}}_{i,j,k}
\end{equation}
In this way, the $k$-th feature descriptor contains the information of $\boldsymbol{F}$ at the activation of the corresponding attention map $\boldsymbol{A}_k$. 

% \cite{xie2018comparator} use information from multiple discriminative local regions (landmarks) and applied another kind of attentional pooling mechanism on feature maps to obtain landmark specific feature descriptors. Instead of computing the cosine similarity between two feature representations, comparator network directly learns set-wise verification to output similarity score. More precisely, after going through a feature extraction network (ResNet), they obtain not only dense feature representation maps but also a set of attention maps by applying $1\times1$ convolution operations on the feature maps, which also named as landmark score maps. Those landmark score maps will be first re-calibrated by normalizing the landmark responses over the images within a template, then output multiple landmark specific feature descriptors by applying image specific weighted average pooling on the feature maps, defined as follow:
%\begin{eqnarray}
%&& A_{n\dots k} = \frac{exp(A_{n\dots k})}{\sum_{nij}exp(A_{nijk})} \\
%&& V_k = \sum_{nij}F_{nij}\odot A_{nijk} \qquad \text{for} \; k \in [1:K+1]
%\end{eqnarray}
%where $A$ denotes landmark score maps, $F$ denotes intermediate feature maps, $k\in[1,K]$ denotes the $k^{\text{th}}$ landmark score maps and $n\in[1,N]$ denotes the $n^{\text{th}}$ image within the image set. The output is $K+1$ (1 global feature descriptor) feature descriptors with each descriptor representing either one of the face landmarks or global information.

\subsubsection{Aggregate}
We conclude our partial \ac{FR} model with the Aggregate module. Since all feature descriptors $\boldsymbol{v}_k$ focus on different regions within $\boldsymbol{F}$ depending on their corresponding attention map $\boldsymbol{A}_k$, a direct aggregation is impossible. Thus, we map every $\boldsymbol{v}_k$ separately into a joint feature space $\boldsymbol{f}_k \in \mathbb{R}^{256}$ utilizing a single fully connected layer each.
%\begin{equation}
%\boldsymbol{f}_k = \boldsymbol{W}_k \boldsymbol{v}_k
%\end{equation}
%where $\boldsymbol{W}_k \in \mathbb{R}^{256 \times 1024}$ denotes the $k$-th trainable weight matrix. 
Note that as every $\boldsymbol{v}_k$ is in a different feature space, the weights are not shared. Since $\boldsymbol{f}_k$ encode identity information likewise, we compute the mean to obtain the final feature vector $\boldsymbol{f} \in \mathbb{R}^{256}$:
\begin{equation}
\boldsymbol{f} = \frac{1}{K} \sum\limits_k \boldsymbol{f}_k
\end{equation}

\subsection{Loss Functions}

To train our model, we apply a weighted sum of three losses $\mathscr{L}$, which are described in the following:
\begin{equation}  
\mathscr{L}=\ \lambda_{\text{wCE}}\mathscr{L}_{\text{wCE}}+\ \lambda_{\text{wDIV}}\mathscr{L}_{\text{wDIV}}+\lambda_{\text{REG}}\mathscr{L}_{\text{REG}}
\end{equation}
with $\lambda_{\text{wCE}}$, $\lambda_{\text{wDIV}}$ and $\lambda_{\text{REG}}$ denoting hyperparameters to balance the losses, and $\mathscr{L}_{\text{REG}}$ is the $L^2$-norm of all trainable weights.

\subsubsection{Weighted Cross-Entropy $\mathscr{L}_{\text{wCE}}$}

To cope with some vectors $\boldsymbol{f}_k$ representing occluded regions and thus being less relevant, we propose a weighted softmax \ac{CE} loss. As usual for \ac{CE} losses, we add a fully connected layer to every feature vector $\boldsymbol{f}_k$ matching the number of classes in our training dataset. In this way, we obtain $K$ \ac{CE} losses $\mathscr{L}_{\text{CE},k}$. To obtain our final weighted \ac{CE} loss, we scale every $\mathscr{L}_{\text{CE},k}$ with its importance $s_k$ as computed in \autoref{eq:s}:
\begin{equation}
\label{eq:LCE}
\mathscr{L}_{\text{wCE}} =  \sum\limits_k s_k \cdot \mathscr{L}_{\text{CE},k}
\end{equation}

In this way, the network learns to emphasize attention maps representing visible face areas while mitigating the influence of attention maps representing occluded regions. Note that since the weights of the last fully connected layers are shared, every $\boldsymbol{f}_k$ is transformed equally, and thereby, we ensure that they encode identity information likewise, i.e.,  lie in the same feature space. Moreover, due to the high number of classes in the training dataset, $\boldsymbol{f}_k$ act as bottleneck layers improving our network's generalization.

\subsubsection{Weighted Diversity Regularizer $\mathscr{L}_{\text{wDIV}}$}

The diversity regularizer's objective is to assure diversity within the attention maps as, without regularization, the network is prone to tend towards using only one attention map or generating $K$ identical attention maps. We apply an adaptation of the diversity regularizer from Xie et al. \cite{xie2018comparator} to penalize the mutual overlap between different attention maps. First, every attention map $\boldsymbol{A}_k$ is self-normalized into a probability distribution $\boldsymbol{P}_k$ using the softmax function:
\begin{equation}
\boldsymbol{P}_{i,j,k} = \frac{\exp\left(\boldsymbol{A}_{i,j,k}\right)}{\sum\limits_{i,j}\exp\left(\boldsymbol{A}_{i,j,k}\right)}
\end{equation}

Next, we compute their pixel-wise maximum of all $\boldsymbol{P}_k$ scaled with their respective $s_k$ and obtain the sum of all pixels. For mutually non-overlapping attention maps, this sum is close to 1, which allows computing the weighted diversity loss $\mathscr{L}_{\text{wDIV}}$ as follows:
\begin{equation}
\label{eq:loss}
\mathscr{L}_{\text{wDIV}} = 1 - \sum\limits_{i,j} \max_k \left(s_k \cdot \boldsymbol{P}_{i,j,k}\right)
\end{equation}
% if only 1 is active we would punish it in the L_CE pretraining --> maps are kinda fixed and in finetuning it does not diverge

\section{Experiments}
\label{sec:experiments}

\subsection{Training Details}
\label{sec:training}

Our training is divided into two steps. First, we pretrain the model on holistic faces for 20 epochs with ADAM optimizer \cite{ADAM} and a batch size of 50. However, instead of using the weighted \ac{CE} as defined in \autoref{eq:LCE}, we average all $\mathscr{L}_{\text{CE},k}$ 
and balance the losses by setting $\lambda_{\text{wCE}}=\lambda_{\text{wDIV}}=1$ and $\lambda_{\text{REG}}=5\cdot10^{-5}$. We start with an initial learning rate of $0.05$ and divide it by $4$ every 6 epochs. As training dataset, we utilize VGG-Face2 \cite{cao2018vggface2}, which comprises 3.3\,M images of 8631 identities. Using the facial landmarks extracted with MTCNN \cite{MTCNN}, we align every face and crop it to a resolution of $160\times160$ pixels. To improve generalization, we augment the faces by changing brightness, contrast and saturation, and perform left-right flipping with a probability of $50\,\%$. Moreover, dropout with $80\,\%$ keep probability is added after $\boldsymbol{v}_k$. 

\begin{figure}[t]
  \centering
\includegraphics[width=\linewidth]{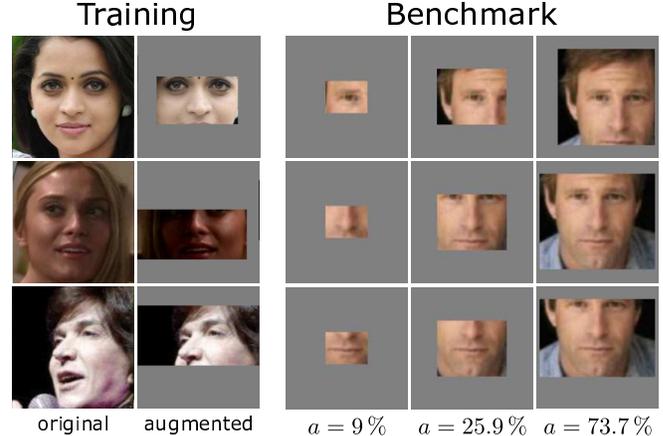}
\vspace{-0.5cm}
  \caption{Faces during finetuning before (first column) and after (second column) applying data augmentation (left). Generation of faces for our centered partial \ac{LFW} protocol for different non-occluded areas $a$ (right).}
  \vspace{-0.4cm}
  \label{fig:training_test}
\end{figure}

In a second step, we leverage that the model performs well on holistic faces and further finetune it on partial faces. As depicted in \autoref{fig:training_test} (left), we synthetically generate rectangular partial faces with an area between $10\,\%$ and $100\,\%$ with a probability of $80\,\%$. Since weights are now well-initialized, we finetune the model for 5 more epochs using an initial learning rate of 0.002 and decay it every 2 epochs. Moreover, we use the weighted \ac{CE} loss $\mathscr{L}_{\text{wCE}}$ as in \autoref{eq:LCE}. All remaining parameters are identical as during pretraining.

\begin{table*}[t]
\begin{small}
  \centering
  \caption{Effect of different parameters on the accuracy in $\%$ on the LFW and CPLFW dataset. Mean accuracies  over nine non-occluded areas $a$ are reported for \textit{partial - holistic} (e.g. mouth - holistic), \textit{partial - same} (e.g. mouth - mouth), and  \textit{partial - cross} (e.g. mouth - nose).}
    \begin{tabular}{ccccc|cccccccc|c}
    \toprule
    &&&&& CPLFW & \multicolumn{7}{c|}{LFW}\\
    \cmidrule(lr){7-13}
          &       &       &       &       &       & &       \multicolumn{3}{c}{non-centered: partial -} &          \multicolumn{3}{c|}{centered: partial - } & \# \\ 				\cmidrule(lr){8-10} \cmidrule(lr){11-13}
   $K$ & Agg & $\mathscr{L}_{\text{wCE}}$ & $f_{\text{ex}}$  & $f_{\text{norm}}$    & holistic & holistic & holistic & same & cross & holistic & same & cross & Params \\
    \midrule
    \multicolumn{5}{l|}{ResNet-41}  & 87.52
& 99.62 & 97.71 & 97.27 & 94.53 & 97.25 & 96.80 & 93.56 & \phantom{0}8.82\,M \\
    \multicolumn{5}{l|}{ResNet-50 (no finetune)} & 88.20 & 99.58 & 94.77 & 94.93 & 88.85 & 92.05 & 92.47 & 83.92 & 24.05\,M \\
    \multicolumn{5}{l|}{ResNet-50}  & 87.80
& 99.60 & 97.75 & 97.36 & 94.80 & 95.48 & 94.72 & 89.60 & 24.05\,M \\
    \midrule
    5     &       &       & \multicolumn{2}{c|}{no re-calibration} & 87.80 & 99.47 & 97.60 & 97.18 & 94.14 & 97.01 & 96.64 & 92.96 & 16.19\,M \\
    5     &       &       & softmax & softmax & 88.42 & 99.45 & 97.76 & 97.30 & 94.23 & 97.25 & 96.77 & 93.00 & 16.19\,M \\
%        5     &       &       & -     & sigmoid & 89.23 & 99.60 & 97.96 & 97.53 & 94.58 & 97.66 & 97.21 & 93.82 & 16.19 \\
    5     &       &       & softmax & sigmoid & 88.87 & 99.62 & \textbf{98.04} & 97.60 & 94.58 & 97.62 & \textbf{97.16} & 93.45 & 16.19\,M \\
    5     & $\surd$ &       & softmax & sigmoid & 89.10 & 99.47 & 98.02 & 97.56 & 94.79 & 97.61 & 97.12 & 93.74 & 17.25\,M \\
    5     & $\surd$ & $\surd$ & softmax & sigmoid & \textbf{89.18} & 99.67 & 97.99 & 97.54 & 94.79 & 97.58 & 97.08 & 93.73 & 17.25\,M \\
    \midrule
    12    &       &       & \multicolumn{2}{c|}{no re-calibration} & 88.03 & 99.63 & 97.74 & 97.28 & 94.38 & 97.17 & 96.63 & 93.06 & 16.20\,M \\
    12    &       &       & softmax & softmax & 88.10 & 99.50 & 97.61 & 97.11 & 94.43 & 96.77 & 96.24 & 92.68 & 16.20\,M \\
%        12    &       &       & -     & sigmoid & 88.83 & 99.65 & 97.98 & 97.48 & 94.61 & 97.60 & 97.07 & 93.71 & 16.20 \\
    12    &       &       & softmax & sigmoid & 89.13 & 99.62 & 97.99 & 97.61 & 94.62 & 97.54 & 97.03 & 93.44 & 16.20\,M \\
    12    & $\surd$ &       & softmax & sigmoid & 89.08 & 99.60 & 98.02 & 97.56 & 94.85 & 97.60 & 97.08 & 93.86 & 19.09\,M \\
    12    & $\surd$ & $\surd$ & softmax & sigmoid & 88.97 & \textbf{99.70} & 98.03 & \textbf{97.66} & \textbf{94.90} & \textbf{97.64} &\textbf{97.16}& \textbf{93.87} & 19.09\,M \\    
    \bottomrule
    \vspace{-0.5cm}
    \end{tabular}%
  \label{tab:results}%
\end{small}
\end{table*}%

\subsection{Benchmark Details}

We evaluate our approach on the synthetically-occluded partial \acf{LFW} dataset \footnote{\url{https://github.com/stefhoer/PartialLFW}}, which is based on \ac{LFW} \cite{LFW}. To generate partial faces, we crop rectangular face patches of nine different areas ranging from $a=9\,\%$ to $a=73.7\,\%$ of the original face around three landmarks: left eye, nose, and mouth. Next, we either leave the cropped face patches at their initial location and fill the remaining image with zeros (\textit{non-centered}) or move the patch to the center and zero-pad to match the input resolution (\textit{centered}). Moreover, we utilize the CPLFW \cite{zheng2018cross} dataset, which contains naturally occurring occlusion due to extreme head poses.  As distance measure, we utilize the cosine distance of the features $\boldsymbol{f}$.

\newlength\figureheight
\newlength\figurewidth
\setlength\figureheight{5.75cm}
\setlength\figurewidth{1.0\columnwidth}

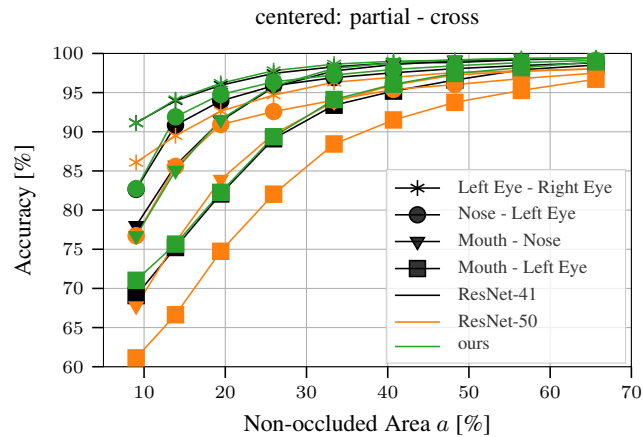
\begin{figure}[b!]
\vspace{-0.5cm}
% This file was created by tikzplotlib v0.9.6.

\pgfplotsset{every tick label/.append style={font=\scriptsize}}
\begin{tikzpicture}

\definecolor{color0}{rgb}{1,0.498039215686275,0.0549019607843137}
\definecolor{color1}{rgb}{0.12156862745098,0.466666666666667,0.705882352941177}
\definecolor{color2}{rgb}{0.172549019607843,0.627450980392157,0.172549019607843}

\begin{axis}[
height=\figureheight,
legend cell align={left},
legend style={fill opacity=0.8,
font=\scriptsize, 
draw opacity=1, text opacity=1, at={(0.99,0.01)}, anchor=south east, draw=white!80!black},
tick align=outside,
ylabel near ticks,
tick pos=left,
title={centered: partial - cross},
width=\figurewidth,
x grid style={white!69.0196078431373!black},
xlabel={Non-occluded Area \(\displaystyle a\) [\%]},
xmajorgrids,
xmin=5, xmax=70,
xtick style={color=black},
y grid style={white!69.0196078431373!black},
ylabel={Accuracy [\%]},
ymajorgrids,
every major tick/.style={black, semithick},
ymin=60, ymax=100,
ytick={60,65,70,75,80,85,90,95,100},
tick align=outside,
tick pos=left,
ytick style={color=black}
]

\addplot [semithick, black, mark=asterisk, mark size=3, mark options={solid}]
table {%
-3 80
-2 90
};
\addlegendentry{Left Eye - Right Eye}
\addplot [semithick, black, mark=*, mark size=3, mark options={solid}]
table {%
-3 80
-2 90
};
\addlegendentry{Nose - Left Eye}
\addplot [semithick, black, mark=triangle*, mark size=3, mark options={solid,rotate=180}]
table {%
-3 80
-2 90
};
\addlegendentry{Mouth - Nose}

% ResNet-41
\addplot [semithick, black, mark=*, mark size=3, mark options={solid}, forget plot]
table {%
9.0234375 82.642
13.87109375 90.842
19.44140625 93.942
25.94921875 95.875
33.39453125 96.917
40.70963542 97.533
48.27083333 98.058
56.45703125 98.458
65.63151042 98.808
};
\addplot [semithick, black, mark=square*, mark size=3, mark options={solid}, forget plot]
table {%
9.0234375 69.083
13.87109375 75.225
19.44140625 82.058
25.94921875 89.108
33.39453125 93.392
40.70963542 95.192
48.27083333 96.608
56.45703125 97.917
65.63151042 98.525
};
\addplot [semithick, black, mark=triangle*, mark size=3, mark options={solid,rotate=180}, forget plot]
table {%
9.0234375 78
13.87109375 85.7
19.44140625 91.392
25.94921875 95.7
33.39453125 97.775
40.70963542 98.583
48.27083333 99.058
56.45703125 99.275
65.63151042 99.4
};
\addplot [semithick, black, mark=asterisk, mark size=3, mark options={solid}, forget plot]
table {%
9.0234375 91.142
13.87109375 93.967
19.44140625 95.942
25.94921875 97.458
33.39453125 98.3
40.70963542 98.792
48.27083333 98.842
56.45703125 99.2
65.63151042 99.392
};

% ResNet-50
\addplot [semithick, color0, mark=*, mark size=3, mark options={solid}, forget plot]
table {%
9.0234375 76.733
13.87109375 85.583
19.44140625 90.925
25.94921875 92.583
33.39453125 94.033
40.70963542 95.408
48.27083333 96.017
56.45703125 96.792
65.63151042 97.5
};
\addplot [semithick, color0, mark=square*, mark size=3, mark options={solid}, forget plot]
table {%
9.0234375 61.133
13.87109375 66.633
19.44140625 74.742
25.94921875 82.033
33.39453125 88.442
40.70963542 91.508
48.27083333 93.758
56.45703125 95.283
65.63151042 96.708
};
\addplot [semithick, color0, mark=triangle*, mark size=3, mark options={solid,rotate=180}, forget plot]
table {%
9.0234375 67.75
13.87109375 75.917
19.44140625 83.9
25.94921875 89.7
33.39453125 93.85
40.70963542 95.975
48.27083333 97.258
56.45703125 97.725
65.63151042 98.025
};
\addplot [semithick, color0, mark=asterisk, mark size=3, mark options={solid}, forget plot]
table {%
9.0234375 86.083
13.87109375 89.492
19.44140625 92.642
25.94921875 94.683
33.39453125 96.325
40.70963542 96.95
48.27083333 97.567
56.45703125 97.767
65.63151042 98.042
};

%\addplot [semithick, color1, mark=*, mark size=3, mark options={solid}, forget plot]
%table {%
%9.0234375 82.8
%13.87109375 91.3
%19.44140625 94.767
%25.94921875 96.217
%33.39453125 97.383
%40.70963542 97.908
%48.27083333 98.483
%56.45703125 98.725
%65.63151042 98.975
%};
\addplot [semithick, color2, mark=*, mark size=3, mark options={solid}, forget plot]
table {%
9.0234375 82.742
13.87109375 91.908
19.44140625 94.758
25.94921875 96.367
33.39453125 97.258
40.70963542 97.958
48.27083333 98.425
56.45703125 98.833
65.63151042 99.117
};
\addplot [semithick, black, mark=square*, mark size=3, mark options={solid}]
table {%
-3 80
-2 90
};
\addlegendentry{Mouth - Left Eye}
%\addplot [semithick, color1, mark=square*, mark size=3, mark options={solid}, forget plot]
%table {%
%9.0234375 67.592
%13.87109375 72.342
%19.44140625 79.975
%25.94921875 88.917
%33.39453125 93.608
%40.70963542 95.958
%48.27083333 97.492
%56.45703125 98.258
%65.63151042 98.767
%};
\addplot [semithick, color2, mark=square*, mark size=3, mark options={solid}, forget plot]
table {%
9.0234375 71.042
13.87109375 75.658
19.44140625 82.267
25.94921875 89.367
33.39453125 94.117
40.70963542 96.067
48.27083333 97.442
56.45703125 98.192
65.63151042 98.933
};

%\addplot [semithick, color1, mark=triangle*, mark size=3, mark options={solid,rotate=180}, forget plot]
%table {%
%9.0234375 75.133
%13.87109375 83.842
%19.44140625 91.35
%25.94921875 95.583
%33.39453125 97.683
%40.70963542 98.683
%48.27083333 99.05
%56.45703125 99.317
%65.63151042 99.308
%};
\addplot [semithick, color2, mark=triangle*, mark size=3, mark options={solid,rotate=180}, forget plot]
table {%
9.0234375 76.7
13.87109375 85.058
19.44140625 91.558
25.94921875 95.833
33.39453125 98.092
40.70963542 98.817
48.27083333 99.183
56.45703125 99.375
65.63151042 99.383
};
\addplot [semithick, black]
table {%
-3 80
-2 90
};
\addlegendentry{ResNet-41}
\addplot [semithick, color0]
table {%
-3 80
-2 90
};
\addlegendentry{ResNet-50}
%\addplot [semithick, color1, mark=asterisk, mark size=3, mark options={solid}, forget plot]
%table {%
%9.0234375 91.283
%13.87109375 93.733
%19.44140625 96.375
%25.94921875 97.958
%33.39453125 98.575
%40.70963542 98.858
%48.27083333 99.042
%56.45703125 99.225
%65.63151042 99.283
%};
%\addplot [semithick, color1]
%table {%
%-3 80
%-2 90
%};
%\addlegendentry{Attend}
\addplot [semithick, color2, mark=asterisk, mark size=3, mark options={solid}, forget plot]
table {%
9.0234375 91.125
13.87109375 94.1
19.44140625 96.192
25.94921875 97.8
33.39453125 98.633
40.70963542 99.008
48.27083333 99.142
56.45703125 99.367
65.63151042 99.442
};
\addplot [semithick, color2]
table {%
-3 80
-2 90
};
\addlegendentry{ours}
\end{axis}

\end{tikzpicture}
\vspace{-0.5cm}
\caption{Accuracy on the LFW dataset for $K=12$, $f_\text{ex}=\sigmoid$, and $f_\text{norm}=\softmax$ with Aggregate and $\mathscr{L}_{\text{wCE}}$ on the centered partial - cross protocol dependent on the non-occluded area $a$.}
\label{fig:cross_dependancy}
\end{figure}

\subsection{Baselines}

We compare our approach with a standard \textit{ResNet-50} and a \textit{ResNet-41}, which is obtained by removing the last block of a \textit{ResNet-50} and, thus, has the same depth as our approach. Both are trained on softmax \ac{CE} loss and with identical parameters as in \autoref{sec:training}.  We also train without the Aggregate module (\textit{Agg}) by averaging all $K$ normalized attention maps $\boldsymbol{\tilde{A}}_k$ to obtain a global attention map. Then, attentional pooling is applied only for the global attention map followed by a single bottleneck layer to obtain the feature vector.

\subsection{Results}

\autoref{tab:results} depicts the aggregated accuracies for different benchmark protocols on the LFW dataset. When considering a \textit{ResNet-50 (no finetune)}, which was never exposed to partial faces during training, we can observe that standard \ac{FR} models are very susceptible to partial faces. By finetuning on partial faces, the model performs better on the partial protocols. While \textit{ResNet-50} outperforms \textit{ResNet-41} on the \textit{non-centered} protocols, it is inferior on the \textit{centered} protocols. We believe that this is due to \textit{ResNet-50} containing more trainable parameters. Thus, it is more prone to overfit on the spatial information present during training, since centering was not part of our data augmentation.

Our ablation study shows that the re-calibration with $f_\text{ex}$\,$=\sigmoid$ and $f_\text{norm}$$\,=\,$$\softmax$ is crucial for a well-performing model. The number of attention maps $K$ only has a minor influence, yet models with $K$$\,=\,$$12$ tend to have superior performance. The Aggregate module and weighted \ac{CE} loss $\mathscr{L}_{\text{wCE}}$ improve the performance, especially in the \textit{partial - cross} protocols. For naturally occurring occlusions as in CPLFW, our model also improves the baseline. Besides, our approach further boosts the accuracy on the \textit{holistic} LFW benchmark, suggesting that our Attend and Aggregate modules combined with partial faces as data augmentation assist generalization. Overall, our approach to partial \ac{FR} outperforms all baselines while comprising fewer parameters than \textit{ResNet-50}.

In \autoref{fig:cross_dependancy}, we illustrate the influence of the non-occluded area $a$ of partial faces in the \textit{centered: partial - cross} protocol. While the accuracy when recognizing \textit{Left Eye - Right Eye} is only slightly affected by $a$, the scenario of verifying whether \textit{Mouth - Left Eye} belong to the same identity is considered most challenging. Overall, we can conclude that our model is more robust compared to the baseline for all \textit{centered: partial - cross} cases.

\section{Conclusion}

In this paper, we propose a \ac{CNN} for partial \ac{FR} consisting of three modules: 1) Extract to predict feature maps and attention maps using a truncated ResNet-50; 2) Attend to re-calibrate the attention maps and perform attentional pooling; 3) Aggregate to fuse the feature information into one global feature vector. 

Our exhaustive analysis demonstrates that our approach outperforms all baselines and provides satisfying results for the arguable more challenging \textit{partial - cross} protocols (e.g. mouth - nose). %, which are not considered in most related works. 
These results suggest that our model successfully transforms any arbitrary face patch into a joint feature space, in which even the matching of  non-overlapping face patches is possible.

% As there is no public available partial face benchmark dataset, we build a standard benchmark dataset for partial face recognition, which contains different scales of cropped faces as well as different cropping facial landmarks. A wide range of experiments show the effectiveness of the proposed attention-based partial face network. We first select a meaningful model ResNet-41 as our baseline model in the experiments. The experiment on full face verification demonstrates that the proposed approach is able to achieve a high accuracy on full face recognition. Then we compare different normalization methods in attend module, different number of attention maps as well as with/without aggregation module on partial face verification. At the end, we demonstrate that our method which uses squeeze-softmax-sigmoid normalization in attend module together with aggregation module outperforms the baseline model ResNet-41. Aggregation has a significant promotion on cross part partial face verification. Moreover, increase the number of attention maps does not guarantee a better performance, it depends on the level of cropping. 


\bibliographystyle{IEEEbib}
\bibliography{ref}

\begin{thebibliography}{10}

\bibitem{Liu.2017}
W.~Liu, Y.~Wen, Z.~Yu, M.~Li, B.~Raj, and L.~Song,
\newblock ``{SphereFace: Deep Hypersphere Embedding for Face Recognition},''
\newblock in {\em {IEEE Conference on Computer Vision and Pattern Recognition
  (CVPR)}}, 2017, pp. 6738--6746.

\bibitem{wang2018cosface}
H.~Wang, Y.~Wang, Z.~Zhou, X.~Ji, D.~Gong, J.~Zhou, Z.~Li, and W.~Liu,
\newblock ``{CosFace: Large Margin Cosine Loss for Deep Face Recognition},''
\newblock in {\em Proceedings of the IEEE Conference on Computer Vision and
  Pattern Recognition (CVPR)}, 2018, pp. 5265--5274.

\bibitem{Arcface}
J.~Deng, J.~Guo, N.~Xue, and S.~Zafeiriou,
\newblock ``{ArcFace: Additive Angular Margin Loss for Deep Face
  Recognition},''
\newblock in {\em Proceedings of the IEEE Conference on Computer Vision and
  Pattern Recognition (CVPR)}, 2019, pp. 4690--4699.

\bibitem{kim2020groupface}
Y.~Kim, W.~Park, M.C. Roh, and J.~Shin,
\newblock ``{GroupFace: Learning Latent Groups and Constructing Group-based
  Representations for Face Recognition},''
\newblock in {\em Proceedings of the IEEE/CVF Conference on Computer Vision and
  Pattern Recognition (CVPR)}, 2020, pp. 5621--5630.

\bibitem{LFW}
E.~G. Huang, G. B. Learned-Miller,
\newblock ``{Labeled Faces in the Wild: Updates and New Reporting
  Procedures},''
\newblock Tech. {R}ep. UM-CS-2014-003, University of Massachusetts, Amherst,
  May 2014.

\bibitem{opitz2016grid}
M.~Opitz, G.~Waltner, G.~Poier, H.~Possegger, and H.~Bischof,
\newblock ``{Grid Loss: Detecting Occluded Faces},''
\newblock in {\em European Conference on Computer Vision (ECCV)}, 2016.

\bibitem{mahbub2016partial}
U.~Mahbub, V.M. Patel, D.~Chandra, B.~Barbello, and R.~Chellappa,
\newblock ``{Partial Face Detection for Continuous Authentication},''
\newblock in {\em IEEE International Conference on Image Processing (ICIP)}.
  IEEE, 2016, pp. 2991--2995.

\bibitem{chen2018adversarial}
Y.~Chen, L.~Song, Y.~Hu, and R.~He,
\newblock ``{Adversarial Occlusion-aware Face Detection},''
\newblock in {\em IEEE 9th International Conference on Biometrics Theory,
  Applications and Systems (BTAS)}. IEEE, 2018, pp. 1--9.

\bibitem{liao2012partial}
S.~Liao, A.~K. Jain, and S.~Z. Li,
\newblock ``{Partial Face Recognition: An Alignment Free Approach},''
\newblock {\em IEEE Transactions on Pattern Analysis and Machine Intelligence},
  vol. 35, no. 5, pp. 1193--1205, 2012.

\bibitem{hu2013robust}
J.~Hu, J.~Lu, and Y.P. Tan,
\newblock ``Robust partial face recognition using instance-to-class distance,''
\newblock in {\em Visual Communications and Image Processing (VCIP)}. IEEE,
  2013, pp. 1--6.

\bibitem{weng2016robust}
R.~Weng, J.~Lu, and Y.P. Tan,
\newblock ``{Robust Point Set Matching for Partial Face Recognition},''
\newblock {\em IEEE Transactions on Image Processing}, vol. 25, no. 3, pp.
  1163--1176, 2016.

\bibitem{he2016multiscale}
L.~He, H.~Li, Q.~Zhang, Z.~Sun, and Z.~He,
\newblock ``{Multiscale Representation for Partial Face Recognition Under Near
  Infrared Illumination},''
\newblock in {\em IEEE 8th International Conference on Biometrics Theory,
  Applications and Systems (BTAS)}. IEEE, 2016, pp. 1--7.

\bibitem{he2018dynamic}
L.~He, H.~Li, Q.~Zhang, and Z.~Sun,
\newblock ``{Dynamic Feature Learning for Partial Face Recognition},''
\newblock in {\em IEEE/CVF Conference on Computer Vision and Pattern
  Recognition (CVPR)}. IEEE, 2018, pp. 7054--7063.

\bibitem{song2019occlusion}
L.~Song, D.~Gong, Z.~Li, C.~Liu, and W.~Liu,
\newblock ``{Occlusion Robust Face Recognition Based on Mask Learning With
  Pairwise Differential Siamese Network},''
\newblock in {\em IEEE/CVF International Conference on Computer Vision (ICCV)}.
  IEEE, 2019, pp. 773--782.

\bibitem{xu2020improving}
X.~Xu, N.~Sarafianos, and I.~A. Kakadiaris,
\newblock ``{On Improving the Generalization of Face Recognition in the
  Presence of Occlusions},''
\newblock in {\em IEEE/CVF Conference on Computer Vision and Pattern
  Recognition Workshops (CVPRW)}, 2020, pp. 798--799.

\bibitem{resnetv2}
K.~He, X.~Zhang, S.~Ren, and J.~Sun,
\newblock ``{Identity Mappings in Deep Residual Networks},''
\newblock in {\em European Conference on Computer Vision (ECCV)}. Springer,
  2016, pp. 630--645.

\bibitem{sato1998partial}
K.~Sato, S.~Shah, and J.~Aggarwal,
\newblock ``{Partial Face Recognition Using Radial Basis Function Networks},''
\newblock in {\em Proceedings Third IEEE International Conference on Automatic
  Face and Gesture Recognition (FG)}. IEEE, 1998, pp. 288--293.

\bibitem{gutta2002investigation}
S.~Gutta, V.~Philomin, and M.~Trajkovic,
\newblock ``{An Investigation into the Use of Partial-Faces for Face
  Recognition},''
\newblock in {\em Proceedings of Fifth IEEE International Conference on
  Automatic Face and Gesture Recognition (FG)}. IEEE, 2002, pp. 33--38.

\bibitem{park2010periocular}
U.~Park, R.~R. Jillela, A.~Ross, and A.~K. Jain,
\newblock ``{Periocular Biometrics in the Visible Spectrum},''
\newblock {\em IEEE Transactions on Information Forensics and Security}, vol.
  6, no. 1, pp. 96--106, 2010.

\bibitem{xie2018comparator}
W.~Xie, L.~Shen, and A.~Zisserman,
\newblock ``Comparator networks,''
\newblock in {\em Proceedings of the European Conference on Computer Vision
  (ECCV)}, 2018, pp. 782--797.

\bibitem{nair2010rectified}
V.~Nair and G.E. Hinton,
\newblock ``{Rectified Linear Units Improve Restricted Boltzmann Machines},''
\newblock in {\em International Conference on Machine Learning (ICML)}, 2010.

\bibitem{hu2018squeeze}
J.~Hu, L.~Shen, and G.~Sun,
\newblock ``{Squeeze-and-Excitation Networks},''
\newblock in {\em IEEE/CVF Conference on Computer Vision and Pattern
  Recognition (CVPR)}, 2018, pp. 7132--7141.

\bibitem{ADAM}
D.~P. Kingma and J.~Ba,
\newblock ``{Adam: {A} Method for Stochastic Optimization},''
\newblock in {\em International Conference in Learning Representations (ICLR)},
  2015.

\bibitem{cao2018vggface2}
Q.~Cao, L.~Shen, W.~Xie, O.M. Parkhi, and A.~Zisserman,
\newblock ``{VGGFace2: A dataset for recognising faces across pose and age},''
\newblock in {\em 13th IEEE International Conference on Automatic Face \&
  Gesture Recognition (FG 2018)}. IEEE, 2018, pp. 67--74.

\bibitem{MTCNN}
K.~Zhang, Z.~Zhang, Z.~Li, and Y.~Qiao,
\newblock ``Joint face detection and alignment using multitask cascaded
  convolutional networks,''
\newblock {\em IEEE Signal Processing Letters}, vol. 23, no. 10, pp.
  1499--1503, 2016.

\bibitem{zheng2018cross}
T.~Zheng and W.~Deng,
\newblock ``{Cross-Pose LFW: A Database for Studying Cross-Pose Face
  Recognition in Unconstrained Environments},''
\newblock {\em Beijing University of Posts and Telecommunications, Tech. Rep},
  vol. 5, 2018.

\end{thebibliography}

\end{document}